\documentclass{article}

\usepackage{hyperref}
\usepackage[accepted]{arxiv2018}
\usepackage{url}
\usepackage{multirow}

\usepackage[utf8]{inputenc} 
\usepackage[T1]{fontenc}    
\usepackage{booktabs}       
\usepackage{nicefrac}       
\usepackage{microtype}      
\usepackage{mathtools}
\usepackage{enumerate}
\usepackage{tabularx}
\usepackage{ragged2e}
\usepackage{titlesec}
\usepackage[tight-spacing=true]{siunitx}
\usepackage{caption}

\titleformat{\paragraph}
{\normalfont\normalsize\bfseries}{\theparagraph}{1em}{}
\titlespacing*{\paragraph}
{0pt}{3.25ex plus 1ex minus .2ex}{1.5ex plus .2ex}

\newcolumntype{C}[1]{>{\Centering}m{#1}}

\usepackage{listings}
\usepackage{amsthm}
\usepackage{amssymb}
\usepackage{framed} 
\usepackage{amsmath}
\usepackage{amssymb}
\usepackage{relsize}
\usepackage{tikz}
\usepackage{xr}
\usepackage{tikz-cd}

\DeclarePairedDelimiter{\norm}{\lVert}{\rVert}
\usepackage{times}

\usepackage{graphicx} 

\usepackage{subfigure} 
\usepackage{wrapfig}

\usepackage{algpseudocode}
\usepackage{algpascal}

\numberwithin{equation}{section}

\usetikzlibrary{backgrounds}
\usetikzlibrary{calc}
\usepackage{soul}
\usepackage{stackrel}
\usepackage{relsize}

\tikzset{fontscale/.style = {font=\relsize{#1}}}

\arxivtitlerunning{Word and Phrase Translation with word2vec}
\title{Word and Phrase Translation with word2vec}

\begin{document}
	
\twocolumn[
	\arxivtitle{Word and Phrase Translation with word2vec}
	
	\arxivsetsymbol{equal}{*}
	
	\begin{arxivauthorlist}
		\arxivauthor{Stefan Jansen }{aai} \\
		\texttt{stefan@applied-ai.com}
	\end{arxivauthorlist}
	
	\arxivcorrespondingauthor{Stefan Jansen}{stefan@applied-ai.com}
	
	\arxivaffiliation{aai}{Applied Artifificial Intelligence}
	
	\arxivkeywords{Machine Learning, arxiv}
	
	\vskip 0.3in
]

\printAffiliationsAndNotice{}

\label{abstract}
\begin{abstract}
Word and phrase tables are key inputs to machine translations, but costly to produce. New unsupervised learning methods represent words and phrases in a high-dimensional vector space, and these monolingual embeddings have been shown to encode syntactic and semantic relationships between language elements. The information captured by these embeddings can be exploited for bilingual translation by learning a transformation matrix that allows matching relative positions across two monolingual vector spaces. This method aims to identify high-quality candidates for word and phrase translation more cost-effectively from unlabeled data.\\
This paper expands the scope of previous attempts of bilingual translation to four languages (English, German, Spanish, and French). It shows how to process the source data, train a neural network to learn the high-dimensional embeddings for individual languages and expands the framework for testing their quality beyond the English language. Furthermore, it shows how to learn bilingual transformation matrices and obtain candidates for word and phrase translation, and assess their quality.
\end{abstract}

\label{introduction}
\section*{Introduction}
A key input for statistical machine translation are bilingual mappings of words and phrases that are created from parallel, i.e., already translated corpora. The creation of such high-quality labeled data is costly, and this cost limits supply given the large number of bilingual language pairs.

\textit{word2vec} \cite{mikolovEfficientEstimationWord2013} is an unsupervised learning method that generates a distributed representation \cite{rumelhartLearningRepresentationsBackpropagating1986} of words and phrases in a shared high-dimensional vector space. \textit{word2vec} uses a neural network that is trained to capture the relation between language elements and the context in which they occur. More specifically, the network learns to predict the neighbors within a given text window for each word in the vocabulary. As a result, the relative locations of language elements in this space reflects co-occurrence in the training text material.

Moreover, this form of representation captures rich information on syntactic and semantic relationships between words and phrases. Syntactic matches like singular and plural nouns, present and past tense, or adjectives and their superlatives can be found through simple vector algebra: the location of the plural form of a word is very likely to be in the same direction and at the same distance as the plural of another word relative to its singular form. The same approach works for semantic relations like, countries and their capitals or family relationship. Most famously, the location of ‘queen’ can be obtained by subtracting ‘man’ from ‘king’ while adding ‘woman’.

Monolingual vector spaces for different languages learnt separately using \textit{word2vec} from comparable text material have been shown to generate similar geometric representations. To the extent that similarities among languages, which aim to encode similar real-word concepts \cite{mikolovEfficientEstimationWord2013}, are reflected in similar relative positions of words and phrases, learning a projection matrix that translates locations between these spaces could help identify matching words and phrases for use in machine translation from mono-lingual corpora only.

The benefit would be a significant expansion of the training material that can be used to produce high-quality inputs for language translation. In addition, the candidates for word and phrase translation identified through this approach can be scored using their distance to the expected location.

The paper proceeds as follows: (1) introduce the word2vec algorithm and the evaluation of its results. (2) outline the learning process for the projection matrix between vector spaces, the corresponding approach to word and phrase translation, and the evaluation of translation quality. (3) describe key steps to obtain and preprocess the Wikipedia input data, and present important descriptive statistics. (4) present empirical results and steps taken to optimize  these results.

\label{method}
\section*{The word2vec Method}
\textit{word2vec} stands in a tradition of learning continuous vectors to represent words \cite{mikolovLinguisticRegularitiesContinuous2013} using neural networks \cite{bengioNeuralProbabilisticLanguage2003}. The \textit{word2vec} approach emerged with the goal to enhance the accuracy of capturing the multiple degrees of similarity along syntactic and semantic dimensions \cite{mikolovDistributedRepresentationsWords2013}, while reducing computational complexity to allow for learning vectors beyond the thus far customary 50-100 dimensions, and for training on more than a few hundred million words \cite{mikolovEfficientEstimationWord2013}.

\subsection*{Feed-Forward Networks}

Feed-forward neural network language models (NNLM) have been popular to learn distributed representations because they outperform Latent Semantic Analysis or Latent Dirichlet Allocation in capturing linear regularities, and in particular the latter in computational cost.

\begin{figure}[ht]
	\centering
	\includegraphics[width=0.47\textwidth]{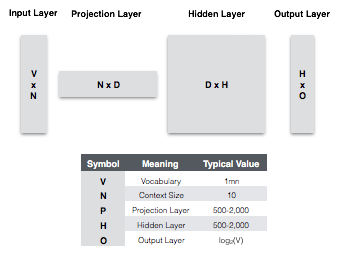}
	\vspace{-0.1in}
	\caption{Feed-Forward NN Architecture}
	\label{fig:ffnn_architecture}
\end{figure}

However, computational cost remained high as NNLM combine input, projection, hidden, and output layer. The input layer contains $N$ neighbors (the input context in 1-of-V encoding) for each word in the vocabulary $V$ to predict a probability distribution over the vocabulary. The context is expanded to a higher dimensionality in the Projection Layer, and then fed forward through a non-linear Hidden Layer. The output probability distribution can be obtained through a Softmax layer, or, more efficiently, a hierarchical Softmax that uses a balanced binary of Huffman tree to reduce output complexity to $\log_2(V)$ or $\log_2(\text{unigram-perplexity}(V))$, respectively \cite{mikolovEfficientEstimationWord2013}.

\subsection*{Recurrent Neural Networks}

Recurrent Neural Networks (RNN) avoid the need to specify the size N of the context, and can represent more varied patterns then NNLM \cite{bengioScalingLearningAlgorithms2007}. RNN have input, hidden and output and no projection layer, but add a recurrent matrix that connects the hidden layer to itself to enable time-delayed effects, or short-term memory. The output layer works as for NNLM.

\begin{figure}[htb]
	\centering
	\includegraphics[width=0.47\textwidth]{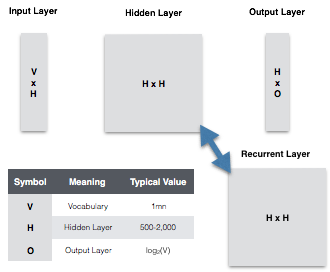}
	\vspace{-0.1in}
	\caption{Recurrent NN Architecture}
	\label{fig:rnn_architecture}
\end{figure}

\subsection*{The Skip-Gram Model}
The computational complexity, or number of parameters corresponds to the matrix multiplications required for back-propagation during the learning process driven by stochastic gradient descent  is large due to the dense, non-linear Hidden Layer. 

\begin{figure*}[htb]
	\centering
	\includegraphics[width=0.97\textwidth]{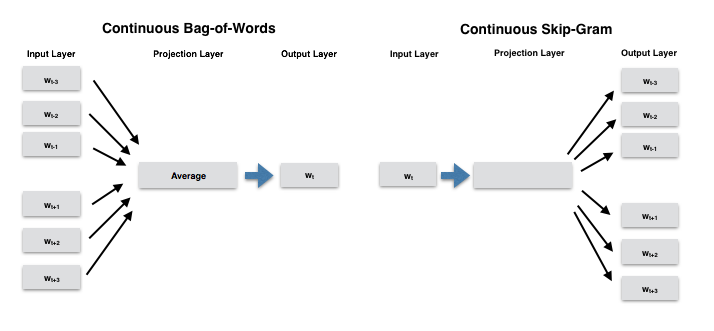}
	\vspace{-0.1in}
	\caption{Continuous Bag of Words \& Skip-Gram Models}
	\label{fig:cbow}
\end{figure*}

The work by \cite{mikolovEfficientEstimationWord2013} has focused on simpler models to learn word vectors, and then train NNLM using these vector representations. The result are two architectures that eliminate the Hidden Layer, and learn word vectors by, as above, predicting a word  using its context, or, alternatively, by predicting the context for each word in the vocabulary.

The \textit{Continuous Bag-of-Word Model} (CBOW) averages the vectors of words in a window before and after the target word for its prediction, as above. The model is called ‘bag of words’ because word order does not matter.

The \textit{Continuous Skip-Gram Model}, in turn changes the direction of the prediction task, and learns word vectors by predicting various individual targets in the context window around each word.
A comparison of these architectures \cite{mikolovEfficientEstimationWord2013} suggests that the Skip-Gram model produces word vectors that better capture the multiple degrees of similarity among words (see below on specific metrics used for this task). For this reason, the experiments in this paper focus on the Skip-Gram Model.

\begin{figure}[htb]
	\centering
	\includegraphics[width=0.47\textwidth]{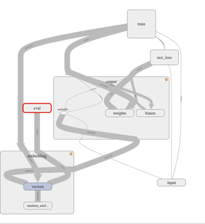}
	\vspace{-0.1in}
	\caption{TensorFlow Computational Graph}
	\label{fig:tf_graph}
\end{figure}

\subsubsection*{Architecture Refinements}
To improve the Skip-Gram model’s accuracy or increase the speed of training, several architecture refinements have been proposed, namely using candidate sampling for a more efficient  formulation of the objective function, and the subsampling of frequently occurring words. 

\paragraph*{Candidate Sampling}
To find word representations that predict surrounding words within a context $c$ with high accuracy, the Skip-Gram model, for a given sequence of words $w_1, w_2, …,w_N$, maximizes the following \textbf{objective} of average log probability
\[
\frac{1}{N}\sum_{n=1}^{N}\sum_{-c\leq j \leq c, j\neq0} \log \text{P}(w_{n+j}|w_n)
\]

over all $ N $ target words and their respective contexts. 
The probability  predicted for any a context word can be based on the inner product of the vector representations of the input and the output candidates $w_I$ and $w_O$, respectively, normalized to conform to the requirements of a probability distribution over all words in the vocabulary of size $ N $ using to the \textbf{Softmax} function:
\[ \text{P}(w_O|w_I)=\frac{\exp{(v_{w_O}^T}v_{w_I}))}{\sum_{w=1}^{W}\exp(v_W^Tv_{w_I})} \]
	
However, the complexity of calculating these probabilities and related gradients for all words becomes prohibitive as the vocabulary grows.

One alternative is the Hierarchical Softmax \cite{bengioNeuralProbabilisticLanguage2003} that reduces the number of computations to $\log_2N$ by representing the output layer as a balanced binary tree. Using Huffman codes to obtain short codes for frequent words further speeds up training.

An alternative to the Hierarchical Softmax function that reduces the number of computations required for inference and back-propagation is \textbf{Noise Contrastive Estimation} (NCE) \cite{gutmannNoisecontrastiveEstimationUnnormalized2012}. Instead of calculating a probability distribution over all possible target words, NCE uses logistic regression to distinguish a target from samples from a noise distribution. NCE approximately maximizes the log probability of the Softmax. \cite{mikolovEfficientEstimationWord2013} simplify NCE by introducing \textbf{Negative Sampling} (NEG), which uses only samples from the noise distribution and obviates the need for the numerical probabilities of the noise distribution itself. 

Either NCE or NEG replace the expression for $ \log\text{P}(w_{n+j}|w_n) $ in the Skip-Gram objective function, and the network is trained using the back-propagation signals resulting from the probabilities predicted for noise samples and actual context words during training.

\cite{mikolovEfficientEstimationWord2013} suggest values of $ k $  in the range of 2-5 for large, and 5-20 for smaller data sets. In addition, using NEC or NEG requires defining the noise distribution, and the authors recommend drawing from the unigram distribution raised to the power $\dfrac{3}{4}$.

\paragraph*{Subsampling Frequent Words}
The frequency distribution of words in large corpora tends to be uneven. In order to address an imbalance of very frequent, but less informative words that can dilute the quality of the word vectors, \cite{mikolovEfficientEstimationWord2013} introduce subsampling of frequent words, and discard each word $ w_i $ in the training set with a probability:
\[ P(w_i)=1-\sqrt{\frac{t}{f(w_i)}} \]
	
where $ t $ is a threshold (recommended at $ 10^{-5} $), and $ f(w_i) $ is the frequency of word $ w_i $. The benefits of this subsampling approach is to significantly curtail the occurrence of words that are more frequent than $ t $ while preserving the frequency ranking of words overall. The authors report improvements in both training speed and accuracy.

\subsubsection*{Hyper Parameter Choices}
Training the Skip-Gram model on text samples requires choices for preprocessing and the setting of model hyper-parameters.   
Hyper-parameters include:
\begin{enumerate}[(i)]
	\item \textbf{Context size}$ C $: increasing $ C $ has been reported to boost accuracy through a larger number of training samples, but also increases training time. Mikholov et al suggest randomizing the size of the context range  for each training sample with probability $ \frac{1}{C} $, where $ C $ is the maximum context size. In practice, for $ C=5 $, this means selecting $ C=1,2,...,5 $ with probability $ 0.2 $ each. 
	\item \textbf{Minimum count} $ w_{\text{min}} $for words to be included in the vocabulary: words that are observed less often than $ M $ times are replaced by a token ‘UNK’ for unknown and all treated alike.
	\item \textbf{Subsampling frequency} $ F_{\text{sub}} $: as mentioned above, recommended at $ 10^{-5} $.
	\item \textbf{Size of negative samples}: 2-5 for large samples, 5-20 for smaller training sets.
	\item \textbf{Embedding size} $ D $: the dimensionality of the word vector increases computational complexity just as increasing the training set size. \cite{mikolovEfficientEstimationWord2013} suggest that conventional vector size choices of 50-100 are too small and report significantly better results for ranges 300-1,000.
	\item \textbf{Epochs to train}: ranges from 3-50 have been reported, but this choice often depends on the constraints imposed by the size of the training set, the computational complexity resulting from other parameter choices, and available resources.
\end{enumerate}

Additional choices include the \textbf{optimization algorithm}. Both standard stochastic gradient and adaptive methods like Adagrad or AdamOptimizer have been used. In either case, the \textbf{initial learning rate} and possibly \textbf{decay rates} may have to be selected, as well as \textbf{batch sizes}.

\subsubsection*{Evaluating Vector Quality}
While the Skip-Gram model is trained to accurately predict context words for any given word, the desired output of the models are the learnt embeddings that represent each word in the vocabulary. 

A number of tests have been developed to assess whether these vectors represent the multiple degrees of similarity that characterize words. These tests are based on the observation that word vectors encode semantic and syntactic relations in the relative locations of words, and that these relations can be recovered through vector algebra. In particular, for a relation $ C:D $ analogous to $ A:B $, the location of $ D $ should closely correspond to following operation: 
\[ \vec{D} = \vec{C} + (\vec{B} - \vec{A})\]
	
\cite{mikolovEfficientEstimationWord2013} have made available over 500 analogy pairs in 15 syntactic and semantic categories. Within each category, the base pairs are combined to yield four-valued analogies. For each, the location $ \vec{D}$ of the fourth word is calculated from the above vector operation. Then, the cosine distance of the correct fourth term is calculated, and compared to other words found in the neighborhood of the projected $ \vec{D} $. According to the  metric $ P@k $used for evaluation, a prediction is deemed correct when the correct term appears among the nearest $ k $ neighbors.

\begin{table}[h]\footnotesize
	\centering
	\caption{Analogy Test}
	\begin{tabular}{llrll}
		\toprule
		{} Topic &  Count &     A or C &       B or D \\
		\midrule
			Capital-common-Countries &     22 &      tokyo &        japan \\
			Capital-World &     92 &     zagreb &      croatia \\
			City-in-State &     66 &  cleveland &         ohio \\
			Currency &     29 &    vietnam &         dong \\
			Family &     22 &      uncle &         aunt \\
			Adjective-to-Adverb &     31 &      usual &      usually \\
			Opposite &     28 &   tasteful &  distasteful \\
			Comparative &     36 &      young &      younger \\
			Superlative &     33 &      young &     youngest \\
			Present-Participle &     32 &      write &      writing \\
			Nationality-Adjective &     40 &    ukraine &    ukrainian \\
			Past-Tense &     39 &    writing &        wrote \\
			Plural &     36 &      woman &        women \\
			Plural-verbs &     29 &      write &       writes \\
		\bottomrule
	\end{tabular}
\end{table}

In order to adapt this test to multiple languages, I translated the base pairs using the google translate API and then manually verified the result.

A few complications arise when translating the English word pairs into German, Spanish and French: 
\begin{itemize}
	\item translation results in a single word in the target language (e.g., some adjectives and adverbs have the same word form in German), rendering the sample unsuitable for the geometric translation test. These pairs were excluded for the affected language.
	\item the translation of the single-word source results in ngrams with $ n_1 $ In these cases, I combined the ngrams to unigrams using underscores and replaced the ngrams with the result.
\end{itemize}

\subsection*{Learning a Projection Matrix}
The result of each monolingual Skip-Gram model is an embedding vector of dimensions $ V_L x D_L $, where $ V_L $ is the size of the vocabulary for the language $ L $, and $ D_L $ is the corresponding embedding size. Hence, for each word $ w_i^S $ in the source language $ S $ there is a vector $ s_i\in\mathbb{R}^{D_S} $, and for each word $w_j^T$ in the target language $ T $,there is a vector $ t_i\in\mathbb{R}^{D_T} $.

We need to find a translation matrix $ W $ so that for a correctly translated pair $ \{w_k^S, w_k^T\} $, the matrix approximately translates the vector locations so that $ W_{S_k}\approx t_k $.
The solution can be find by solving the optimization problem
\[ \min_W \sum_{k=1}^{N} \norm{W s_k - t_k}^2 \]
using gradient descent to minimize the above $ L^2 $ loss.
The resulting translation matrix $ \hat{W} $ will estimate the expected location of a matching translation for any word in the source language provided its vector representation, and use cosine distance to identify the nearest candidates.
In practice, the translation matrix will be estimated using known translations obtained via google translate API.

\label{data}
\section*{The Data: Wikipedia}
\subsection*{Corpus Size \& Diversity}

The empirical application of \textit{word2vec} word and phrase translation uses monolingual Wikipedia corpora available \href{https://dumps.wikimedia.org/}{online} in the four languages shown in Table~\ref{tab:languages}. 

\begin{figure}[htb]
	\centering
	\includegraphics[width=0.47\textwidth]{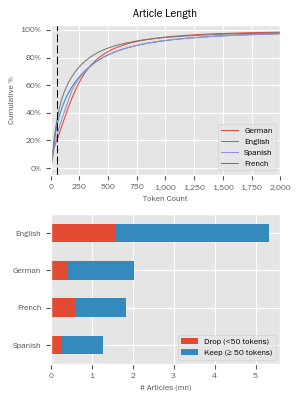}
	\vspace{-0.1in}
	\caption{Article Length Distribution}
	\label{fig:article_length}
\end{figure}

The English corpus is by over two times larger than the second largest German corpus, counting 5.3 million articles, over 2 billion tokens and 8.2 million distinct word forms. 

\begin{table}[!ht]\footnotesize
	\centering
	\caption{Wikipedia corpus statistics (in million)}\label{tab:languages}
	\begin{tabular}{lrrrr}
		\toprule
		Language &  Articles &  Sentences &  Tokens &  Word Forms \\
		\midrule
		English &       5.3 &       95.8 &  2091.5 &         8.2 \\
		German &       2.0 &       49.8 &   738.1 &         8.9 \\
		Spanish &       1.3 &       21.2 &   637.1 &         3.0 \\
		French &       1.8 &       20.9 &   571.4 &         3.1 \\
		\bottomrule
	\end{tabular}
\end{table}

High-level statistics for the four language corpora highlight some differences between languages that may impact translation performance. For instance, the German corpus contains more unique word forms than the English corpus while only containing 35\% of the number of tokens. The number of tokens per unique word form are 85 for German, 187 for French, 216 for Spanish and 254 for English.

In order to exclude less meaningful material like redirects and others, articles with fewer than 50 words were excluded, resulting in the reduced sample sizes shown in Figure~\ref{fig:article_length}.

\subsection*{Parts-of-Speech \& Entity Tags}

To obtain further insight into differences in language structure, the corpora were parsed using \href{https://spacy.io/}{SpaCy} (English \& German) and 
\href{https://stanfordnlp.github.io/CoreNLP/}{Stanford CoreNLP} (French \& Spanish).

\begin{figure}[htb]
	\centering
	\includegraphics[width=0.47\textwidth]{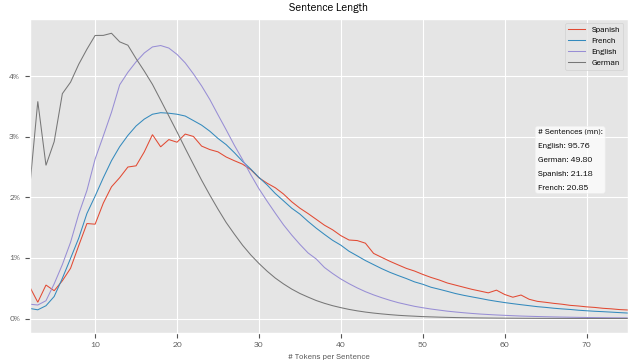}
	\vspace{-0.1in}
	\caption{Sentence Length Distribution}
	\label{fig:sentence_length}
\end{figure}

Sentence parsing revealed markedly shorter sentences in German, compared to longer sentences in Spanish and French (see Figure \ref{fig:sentence_length}).

\subsubsection*{Phrase Modeling}

I used a simple approach to identify bigrams that represent common phrases. In particular, I used to following scoring formula to identify pairs of words that occur more likely together:

\[
\text{score}(w_i, w_j)=\frac{\text{count}(w_i, w_j) - \delta}{\text{count}(w_i)\text{count}(w_j)}
\]
	
where $ \delta $ represents a minimum count threshold for each unigram.

\begin{table}[ht]
	\centering
	\caption{Most frequent words}
	\label{tab:freq_words}
	\begin{tabular}{cllll}
		\toprule
		Rank & English & German & Spanish & French \\
		\midrule
		1 &     the &      . &      de &      , \\
		2 &       , &      , &       , &     de \\
		3 &       . &    der &      el &      . \\
		4 &      of &    die &      la &     la \\
		5 &     and &    und &       . &     le \\
		6 &      in &     in &      en &      " \\
		7 &       " &     `` &       y &     et \\
		8 &      to &    von &       a &     l' \\
		9 &       a &      ) &       " &      à \\
		10 &     was &    den &     que &    les \\
		\bottomrule
	\end{tabular}
\end{table}

To allow for phrases composed of more than two unigrams, bigrams were scored repeatedly, after each iteration combined if their score exceeded a threshold. The threshold was gradually reduced. After three iteration, about 20\% of the tokens consisted of ngrams.

\begin{table}[ht]\footnotesize
	\caption{Most frequent ngrams}\label{tab:ngrams}
	\begin{tabular}{llll}
		\toprule
		English &         German &         Spanish &       French \\
		\midrule
		such as &      vor allem &     sin embargo &       c' est \\
		has been &          z. b. &  estados unidos &       qu' il \\
		as well as &        gibt es &        así como &     à partir \\
		United States &        im jahr &    se encuentra &    ainsi que \\
		had been &  unter anderem &     por ejemplo &  par exemple \\
		have been &     nicht mehr &      cuenta con &        a été \\
		based on &  befindet sich &       junto con &     au cours \\
		known as &       im jahre &   se encuentran &         n' a \\
		would be &         gab es &        lo largo &   la plupart \\
		New York &   zum beispiel &    se convirtió &     au début \\
		\bottomrule
	\end{tabular}
\end{table}

\label{results}
\section*{Results}

\subsection*{Hyperparameter Tuning}

Since model training with the entire corpus is quite resource intensive (90 min per epoch for the English corpus on 16 cores), we tested various model configurations and preprocessing techniques on smaller subsets of the data comprising of 100 million tokens each.

The models used text input at various stages of preprocessing:

\begin{itemize}
	\item raw: unprocessed text
	\item clean: removed punctuation and symbols recognized by language parsers
	\item ngrams: identified by phrase modeling process
\end{itemize}

The following hyper parameters were tested (baseline in bold):
\begin{itemize}
\item $ D $: Embedding Dimensionality (100, \textbf{200}, 300)
\item $ k $: NCE candidate sample size (10, \textbf{25}, 50)
\item $ t $: subsample threshold (\textbf{1e-3} vs custom threshold, set to subsample 0.2\% most frequent words)
\item $ C $: con\item text window size (3, \textbf{5}, 8)
\end{itemize}

All models were trained for 5 epochs using stochastic gradient descent with a starting learning rate to 0.25 that linearly declined to 0.001. Models were evaluated using Precision@1 on the analogy test using questions that were covered by the reduced sample vocabulary.

The results show:
\begin{itemize}
	\item English language performs significantly better for same (reduced) training set size.
	\item the Spanish and French languages did not benefit from text preprocessing, and achieved their highest score with the raw text input.
	\item German alone performed best with a larger embedding size
	\item English performed best with a larger negative sample size.
\end{itemize}

Relative to baseline values for the same input:
\begin{itemize}
	\item A higher embedding size benefited all languages except English, and reducing the embedding size had a negative impact for all languages.
	\item A larger context improved performance for French and Spanish, while all languages did worse for a smaller context.
	\item A larger number of negative samples had a small positive impact for all except French, while a smaller number reduced performance for all.
	\item A custom (higher) subsampling threshold had a negative impact for English and French and only a small positive impact for German.
\end{itemize}

The validity of these results is limited by the number of training rounds and reduced vocabulary, but provide orientation regarding preferable parameter settings for training with the full set. 

\begin{table*}[ht]
	\centering
	\caption{Sample Results}
	\label{tab:sample_results}
	\begin{tabular}{l *{8}{p{1.4cm}}}
		\toprule
		& \multicolumn{4}{c}{Parameter Settings}&\multicolumn{4}{c}{Precision@1 Analogy Test Results}\\
		Input &  Embedding Size ($ D $) &  Negative Samples ($ k $) & Subsample Threshold ($ t $) &  Context ($ C $) &  English &  French &  Spanish &  German \\
		\midrule
		clean &                      200 &                    50 &                   0.001 &                   5 &   \textbf{0.0411} &  0.0125 &   0.0106 &  0.0070 \\
		clean &                      200 &                    25 &                   0.001 &                   5 &   0.0401 &  0.0124 &   0.0098 &  0.0062 \\
		clean &                      200 &                    25 &                   0.001 &                   8 &   0.0354 &  0.0110 &   0.0109 &  0.0051 \\
		clean &                      300 &                    25 &                   0.001 &                   5 &   0.0345 &  0.0126 &   0.0104 &  \textbf{0.0117} \\
		raw &                      200 &                    25 &                   0.001 &                   5 &   0.0319 &  \textbf{0.0137} &   \textbf{0.0133} &  0.0079 \\
		clean &                      200 &                    25 &                   0.001 &                   3 &   0.0287 &  0.0097 &   0.0101 &  0.0064 \\
		clean &                      100 &                    25 &                   0.001 &                   5 &   0.0263 &  0.0085 &   0.0085 &  0.0055 \\
		clean &                      200 &                    25 &                 custom* &                   5 &   0.0222 &  0.0095 &   0.0098 &  0.0068 \\
		clean &                      200 &                    10 &                   0.001 &                   5 &   0.0186 &  0.0095 &   0.0080 &  0.0059 \\
		ngrams &                      200 &                    25 &                   0.001 &                   5 &   0.0170 &  0.0072 &   0.0078 &  0.0045 \\
		clean, ngrams &                      200 &                    25 &                   0.001 &                   5 &   0.0122 &  0.0086 &   0.0078 &  0.0050 \\
		\bottomrule
	\end{tabular}
\end{table*}

\subsection*{Monolingual Benchmarks}

\cite{mikolovEfficientEstimationWord2013} report the following $ P@1 $ accuracy for the \textit{word2vec} Skip-Gram model in English:

\begin{table}[tb]
	\centering
	\caption{English Benchmark}
	\label{tab:benchmark}
	\begin{tabular}{rlrr}
		\toprule
		Vector Size & \# Words &  Epochs &  Accuracy \\
		\midrule
		300 &    783M &       3 &     36.1\% \\
		300 &    783M &       1 &     49.2\%  \\
		300 &    1.6B &       1 &     53.8\% \\
		600 &    783M &       1 &     55.5\% \\
		1000 &      6B &       1 &    66.5\% \\
		\bottomrule
	\end{tabular}
\end{table}

\subsection{Monolingual \textit{word2vec}}
For the complete Wikipedia corpus for each language, various input formats and hyper parameter were applied to improve results and test for robustness. The choices took into account the sample results from above while adjusting for the larger data set (e.g., a smaller number of negative samples is recommended for larger data).

\begin{table*}[ht]
	\centering
	\caption{Monolingual \textit{word2vec} Model Results}
	\label{tab:mono_results}
	\begin{tabular}{l *{11}{p{1cm}}}
		\toprule
		Model & Loss & Ngrams &  Min. Count &  Vocab Size &  Testable Analogies &  Embed. Size &  Initial LR &  Negative Samples &  Sub-sample Threshold &    P@1 &    P@5 \\
		\midrule
		\multirow{ 7}{*}{English} &  NCE &      y &          25 &      481955 &               25392 &               250 &       0.025 &                25 &                0.0010 &  0.365 &  0.637 \\
		&  NCE &      n &          25 &      481732 &               25392 &               250 &       0.025 &                25 &                0.0010 &  0.378 &  0.626 \\
		&  NCE &      y &         100 &      219341 &               25268 &               200 &       0.030 &                15 &                0.0005 &  0.387 &  0.705 \\
		&  NCE &      y &         500 &       82020 &               23736 &               200 &       0.030 &                15 &                0.0010 &  0.426 &  0.747 \\
		&  NCE &      y &         500 &       82020 &               23736 &               200 &       0.030 &                15 &                0.0010 &  0.434 &  0.878 \\
		&  NEG &      y &         500 &       82020 &               23736 &               200 &       0.030 &                15 &                0.0010 &  0.435 &  0.878 \\
		&  NCE &      y &         500 &       82020 &               23736 &               200 &       0.030 &                15 &                0.0005 &  0.435 &  0.748 \\
		\multirow{5}{*}{French} &  NCE &      y &          25 &      241742 &               18498 &               250 &       0.025 &                25 &                0.0010 &  0.094 &  0.637 \\
		&  NCE &      n &           5 &      624810 &               18832 &               200 &       0.030 &                15 &                0.0010 &  0.100 &  0.626 \\
		&  NCE &      n &          25 &      237069 &               20962 &               250 &       0.030 &                15 &                0.0005 &  0.102 &  0.328 \\
		&  NCE &      y &         250 &       62349 &               13712 &               200 &       0.030 &                15 &                0.0010 &  0.140 &  0.328 \\
		&  NCE &      y &         500 &       43610 &                9438 &               200 &       0.030 &                15 &                0.0010 &  0.190 &  0.328 \\
		\multirow{3}{*}{German} &  NCE &      n &          25 &      536947 &                6414 &               250 &       0.025 &                25 &                0.0005 &  0.119 &  0.253 \\
		&  NCE &      y &         250 &      103317 &                3772 &               200 &       0.030 &                15 &                0.0010 &  0.136 &  0.328 \\
		&  NCE &      y &         500 &       65179 &                2528 &               200 &       0.030 &                15 &                0.0010 &  0.168 &  0.422 \\
		\multirow{3}{*}{Spanish} &  NCE &      n &           5 &      626183 &                8162 &               200 &       0.025 &                25 &                0.0010 &  0.098 &  0.637 \\
		&  NCE &      n &          25 &      236939 &                5026 &               250 &       0.030 &                15 &                0.0010 &  0.106 &  0.626 \\
		&  NCE &      y &         250 &       63308 &                2572 &               200 &       0.030 &                15 &                0.0005 &  0.210 &  0.328 \\
		\bottomrule
	\end{tabular}
\end{table*}

Models with both raw input and phrase (ngram) input were tested using Noise Contrastive Estimation and Negative Sampling, but the latter did not change results (as expected). Embeddings ranged from $ 200-250 $, negative samples from $ 15-25 $, and the initial learning rate from 0.025-0.03, decaying linearly to $ 0.001 $. The subsample threshold varied between $ 5e-4 $ and $ 1e-3 $.

The most significant impact resulted from reduced vocabulary size by increasing the minimum count for a world to be included from 5 for the smaller samples to at least 25, up to 500 (reducing the number of analogies available in the vocabulary, esp. in German and Spanish). 

\begin{figure*}[!h]
	\centering
	\includegraphics[width=0.97\textwidth]{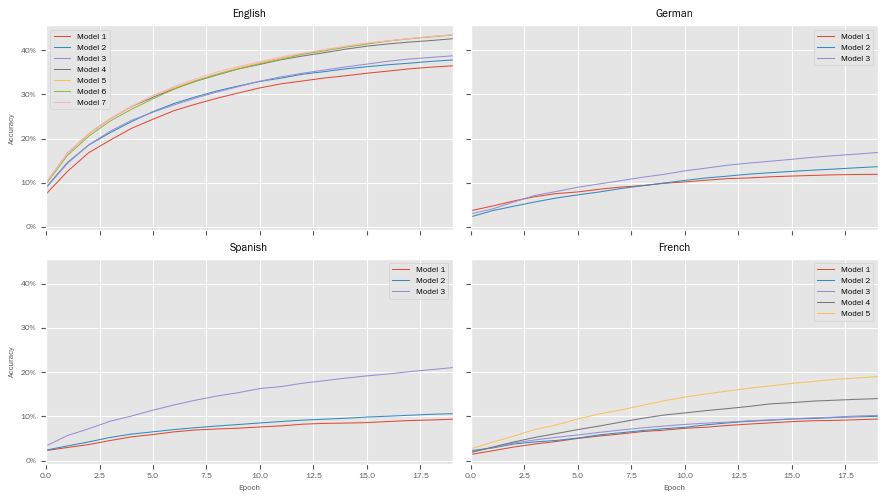}
	\vspace{-0.1in}
	\caption{Model Accuracy}
	\label{fig:model_accuracy}
\end{figure*}

All models were trained for 20 epochs using stochastic gradient descent in TensorFlow.

English, with over 2bn test tokens, performed best (while also covering most analogy word pairs), but did not achieve the benchmark performance reported above. The best models achieved $ P@1 $ accuracy of 43.5\%, and $ P@5 $ accuracy of 87.8\% when the vocabulary was reduced to $ 82,000 $ words. For a vocabulary almost six times this size, $ P@1 $ accuracy was still 37.8\%.

Spanish, French, and German performed less well on the analogy test (which is tailored to the English language). $ P@1 $ accuracy ranges from 16.8\% to 21\%, and $ P@5 $ accuracy from 42.2\% to 63.7\%.

Overall, results behaved as expected in response to changes in parameter settings, with smaller vocabularies boosting precision. The accuracy curves below show that performance was still improving after 20 epochs, so further training (\cite{mikolovEfficientEstimationWord2013} suggest 3-50 epochs) would likely have further increased performance.

\subsection*{Performance across Analogies}
Performance across the various analogy categories is uneven, and a closer look reveals some weaknesses of the models. 
Most models perform strong on capital-country relations for common countries, nationality adjectives, and family relationships. 

\begin{figure*}[!h]
	\centering
		\includegraphics[width=0.97\textwidth]{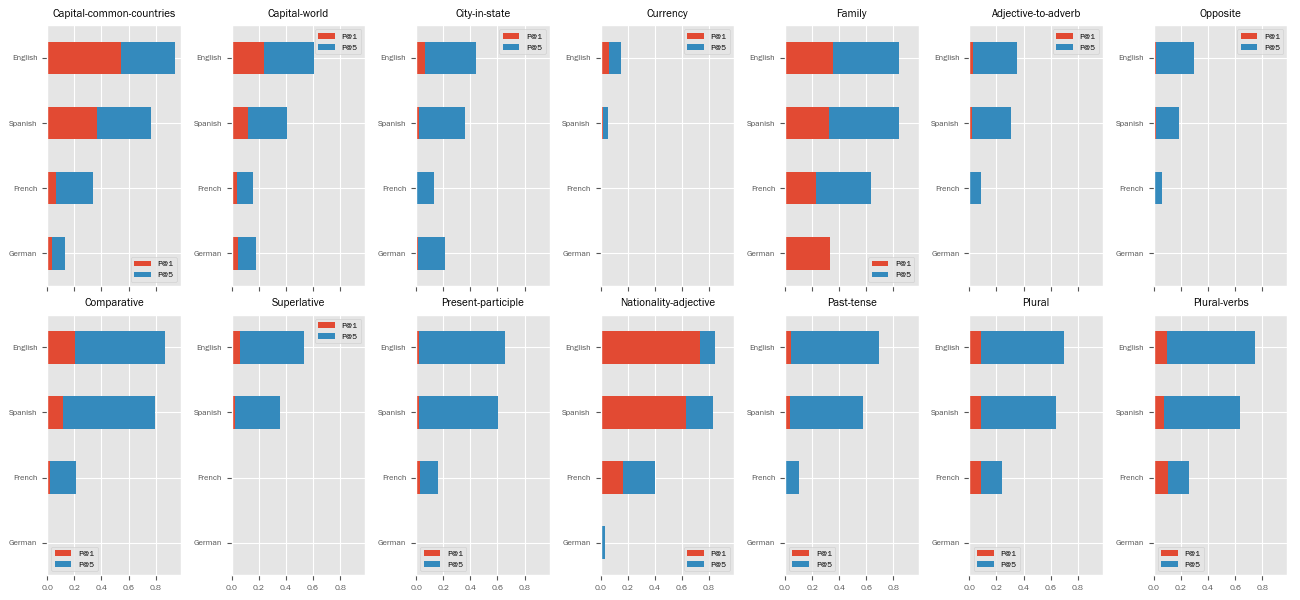}
	\vspace{-0.1in}
	\caption{Model Accuracy by Analogy Topic}
	\label{fig:accuracy_by_topic}
\end{figure*}

Most other areas offer more alternative matches or synonyms, and performance is often strong at the $ P@5 $ level, but not necessarily at the $ P@1 $ level. 
Currency-country relationships are poorly captured for all languages, which might be due to the input text containing fewer information on these (the original tests were conducted on Google news corpora with arguably more economic content)

\subsection*{Translation}

\cite{mikolovExploitingSimilaritiesLanguages2013} report introduced \textit{word2vec} for bilingual translation, and tested English to Spanish translation using the \textit{WMT11} corpus. It provides 575m tokens and 127K unique words in English, and 84m tokens and 107K unique words. The authors reported $ P@1 $ accuracy of 33\%, and $ P@5 $ accuracy of 51\%. 

$ P@5 $ accuracy arguably is a better measure given that there are often several valid translation, while the dictionary contains only one solution (the google translate API does not provides multiple options).

I used the best performing models to test the translation quality from English to any of the other languages, learning the transformation matrix on a training set of the 5,000 most frequent words with matching translations in the process. Results on the test set of 2,500 words per language were best for Spanish with $ P@1 $ at 47.2\% and $ P@5 $ at 62.7\%. French and German performed less well.

\begin{table}[htb]
	\centering
	\caption{$ P@k $ Tranlsation Test Performance}
	\label{tab:test_accuracy}
	\begin{tabular}{rlll}
		\toprule
		P@k & Spanish & French & German \\
		\midrule
		1 &   47.2\% &  33.4\% &  32.0\% \\
		2 &   55.0\% &  39.6\% &  39.3\% \\
		3 &   58.3\% &  43.3\% &  43.9\% \\
		4 &   61.0\% &  45.6\% &  46.4\% \\
		5 &   62.7\% &  47.5\% &  48.3\% \\
		\bottomrule
	\end{tabular}
\end{table}

In order to illustrate how word2vec produces many useful related terms, I am displaying the top 5 option for the English words complexity, pleasure and monsters, marking the dictionary entry provided by google translated in bold.

\begin{table}[!htb]
	\centering
	\caption{Translations for `complexity`}
	\label{tab:complexity}
	\begin{tabular}{lll}
		\toprule
		Spanish &          French &       German \\
		\midrule
		complejidad &      complexité &  Komplexität \\
		predicción &  généralisation &      Dynamik \\
		mecánica cuántica &     formulation &    Präzision \\
		generalización &    probabilités &   dynamische \\
		simplicidad &    modélisation &  dynamischen \\
		\bottomrule
	\end{tabular}
\end{table}

All languages translate complexity correctly, but produces additional interesting associations: both Spanish and French relate complexity to generalization, with Spanish also emphasizing prediction. German closely relates complexity to dynamics.

Pleasure, correctly translated in Spanish and German but with a debatable mistake - according to google translate - in French, also highlights interesting nuances - Spanish and French stress curiosity, while Spanish and German mention patience.
\begin{table}[!htb]
	\centering
	\caption{Translations for `Pleasure`}
	\label{tab:pleasure}
	\begin{tabular}{lll}
		\toprule
		Spanish &      French &     German \\
		\midrule
		placer &    plaisirs &  vergnügen \\
		encanto &   curiosité &    neugier \\
		paciencia &     bonheur &     geduld \\
		curiosidad &  volontiers &    stillen \\
		ternura &     plaisir &   erlebnis \\
		\bottomrule
	\end{tabular}
\end{table}

\begin{figure*}[!h]
	\centering
	\includegraphics[width=0.97\textwidth]{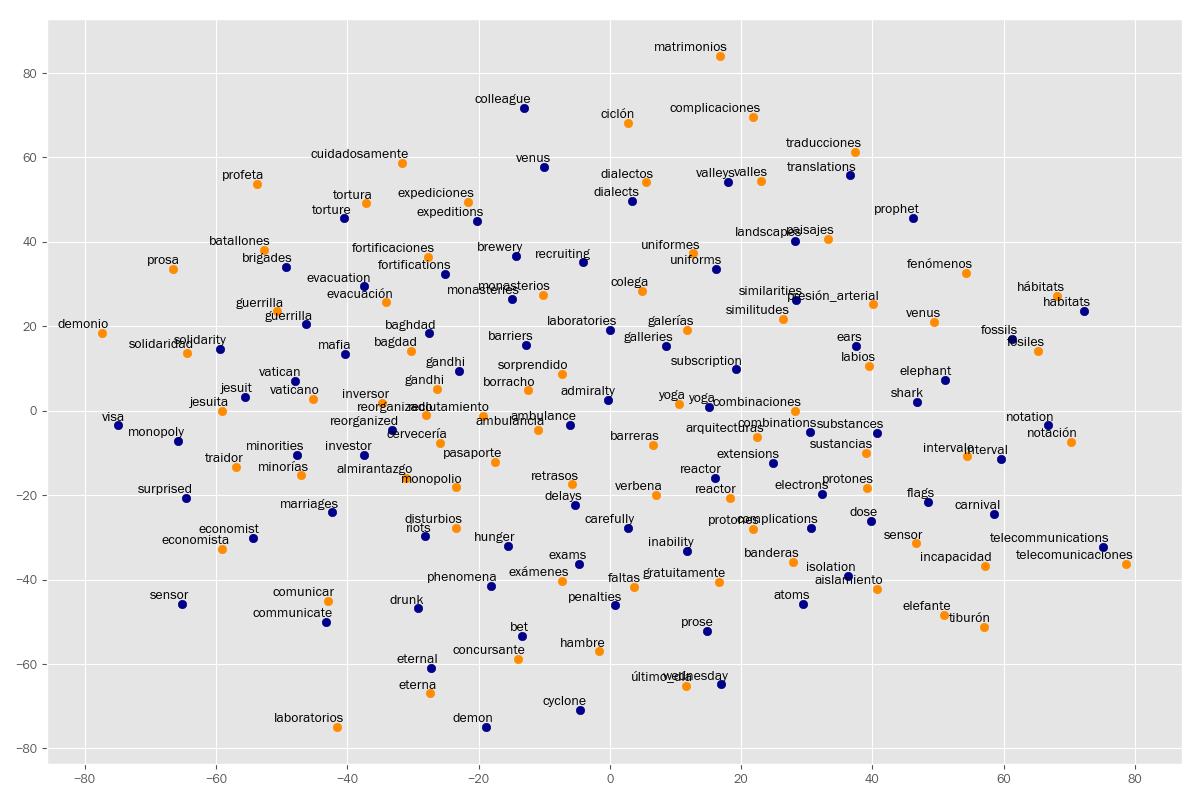}
	\caption{t-SNE Projection of Bilingual \textit{word2vec} Space}
	\label{fig:tsne}
\end{figure*}

Figure \ref{fig:tsne} illustrate the projection of English word vectors into the Spanish vector space using the translation matrix. 

A random sample of 75 English terms  was selected based on correct translation at the $ P@5 $ accuracy level. 
The English vectors were translated into the foreign language space, and both the English vector and the nearest Spanish neighbor vector were projected into two dimensions using \textit{t-distributed stochastic neighbor embedding} (T-SNE) and and displayed with their labels. 

The proximity of similar concepts  both within a language and across language boundaries is striking (see the highlighted shark and elephant example and their translation).

\label{conclusions}
\section*{Conclusions}

It is possible to produce high-quality translations using largely unsupervised learning, with limited supervised learning input to translate between monolingual vector spaces. Preprocessing is also quite limited compared to standard dependency parsing.
 
Using the Wikipedia corpus instead of the smaller WMT11 corpus shows that accuracy can be increased over the previously achieved benchmark. In addition, the translations come with a similarity metric that expresses confidence.

These translations can be extended to multiple bilingual pairs without major effort (e.g. to all six pairs among the four languages considered, and in either direction for 12 applications.

There are multiple opportunities to further improve results:
\begin{itemize}
	\item Monolingual models did not perform optimally, and fell short of benchmark performance. This seems in part to be due to the nature of the analogies chosen for testing, as well as the text material and the quality of the translations. The ‘city-in-state’ pairs emphasize US geography, and the ‘country-currency’ pairs found few matches for all languages, arguably because these terms are used less frequently in Wikipedia than in the Google News corpus used by the original authors. It would certainly be useful to blend multiple corpora to obtain more comprehensive coverage.
	\item The reduced coverage of translated analogies also suggests reviewing in more detail how to adapt the translations. The manual review of the google translation API results produced a significant amount of refinements. Syntactic analogies may also need to be tailored to each language’s grammar as several areas (e.g. adjective-adverb) produced far fewer meaningful pairs in German than in English).
	\item Monolingual models would also benefit from longer training and more computing resources as performance on the analogy test was still improving after the allotted amount of time ended. This would also allow for better hyper parameter settings, in particular with respect to larger context windows (resulting in more training examples per epoch), and higher embedding dimensionality (increasing computational complexity). Furthermore, additional preprocessing options could be explored, e.g. to utilize information from parts-of-speech tagging to distinguish between identical words used in different functions (as noun and verb).
	\item Bilingual models performed quite well despite some shortcomings of the monolingual models, in particular in the English-Spanish case. A thorough review of the translation quality would likely improve the accuracy of the translation matrix. 
	\item The word2vec based translations could possibly be improved by complementing it with additional techniques that have been used successfully, e.g. to remove named entities or use the edit distance to account for morphological similarity.
\end{itemize}

\bibliographystyle{arxiv2018}
\bibliography{bibliography}

\begin{thebibliography}{8}
\providecommand{\natexlab}[1]{#1}
\providecommand{\url}[1]{\texttt{#1}}
\expandafter\ifx\csname urlstyle\endcsname\relax
  \providecommand{\doi}[1]{doi: #1}\else
  \providecommand{\doi}{doi: \begingroup \urlstyle{rm}\Url}\fi

\bibitem[Bengio \& Lecun(2007)Bengio and
  Lecun]{bengioScalingLearningAlgorithms2007}
Bengio, Yoshua and Lecun, Yann.
\newblock {Scaling learning algorithms towards AI}.
\newblock \emph{Large-scale kernel machines}, 2007.
\newblock 00804.

\bibitem[Bengio et~al.(2003)Bengio, Ducharme, Vincent, and
  Jauvin]{bengioNeuralProbabilisticLanguage2003}
Bengio, Yoshua, Ducharme, R{\'e}jean, Vincent, Pascal, and Jauvin, Christian.
\newblock A {{Neural Probabilistic Language Model}}.
\newblock \emph{Journal of Machine Learning Research}, 3\penalty0
  (Feb):\penalty0 1137--1155, 2003.
\newblock ISSN ISSN 1533-7928.
\newblock 03305.

\bibitem[Gutmann \& Hyv{\"a}rinen(2012)Gutmann and
  Hyv{\"a}rinen]{gutmannNoisecontrastiveEstimationUnnormalized2012}
Gutmann, Michael~U. and Hyv{\"a}rinen, Aapo.
\newblock Noise-contrastive {{Estimation}} of {{Unnormalized Statistical
  Models}}, with {{Applications}} to {{Natural Image Statistics}}.
\newblock \emph{J. Mach. Learn. Res.}, 13\penalty0 (1):\penalty0 307--361,
  February 2012.
\newblock ISSN 1532-4435.
\newblock 00247.

\bibitem[Mikolov et~al.(2013{\natexlab{a}})Mikolov, Chen, Corrado, and
  Dean]{mikolovEfficientEstimationWord2013}
Mikolov, Tomas, Chen, Kai, Corrado, Greg, and Dean, Jeffrey.
\newblock Efficient {{Estimation}} of {{Word Representations}} in {{Vector
  Space}}.
\newblock \emph{arXiv:1301.3781 [cs]}, January 2013{\natexlab{a}}.
\newblock 05984.

\bibitem[Mikolov et~al.(2013{\natexlab{b}})Mikolov, Le, and
  Sutskever]{mikolovExploitingSimilaritiesLanguages2013}
Mikolov, Tomas, Le, Quoc~V., and Sutskever, Ilya.
\newblock Exploiting {{Similarities}} among {{Languages}} for {{Machine
  Translation}}.
\newblock \emph{arXiv:1309.4168 [cs]}, September 2013{\natexlab{b}}.
\newblock 00412.

\bibitem[Mikolov et~al.(2013{\natexlab{c}})Mikolov, Sutskever, Chen, Corrado,
  and Dean]{mikolovDistributedRepresentationsWords2013}
Mikolov, Tomas, Sutskever, Ilya, Chen, Kai, Corrado, Greg, and Dean, Jeffrey.
\newblock Distributed {{Representations}} of {{Words}} and {{Phrases}} and
  {{Their Compositionality}}.
\newblock In \emph{Proceedings of the 26th {{International Conference}} on
  {{Neural Information Processing Systems}} - {{Volume}} 2}, NIPS'13, pp.\
  3111--3119, USA, 2013{\natexlab{c}}. {Curran Associates Inc.}
\newblock 07250.

\bibitem[Mikolov et~al.(2013{\natexlab{d}})Mikolov, Yih, and
  Zweig]{mikolovLinguisticRegularitiesContinuous2013}
Mikolov, Tomas, Yih, Wen-tau, and Zweig, Geoffrey.
\newblock Linguistic {{Regularities}} in {{Continuous Space Word
  Representations}}.
\newblock In \emph{Proceedings of the 2013 {{Conference}} of the {{North
  American Chapter}} of the {{Association}} for {{Computational Linguistics}}:
  {{Human Language Technologies}}}, pp.\  746--751, Atlanta, Georgia, June
  2013{\natexlab{d}}. {Association for Computational Linguistics}.
\newblock 01449.

\bibitem[Rumelhart et~al.(1986)Rumelhart, Hinton, and
  Williams]{rumelhartLearningRepresentationsBackpropagating1986}
Rumelhart, David~E., Hinton, Geoffrey~E., and Williams, Ronald~J.
\newblock Learning representations by back-propagating errors.
\newblock \emph{Nature}, 323:\penalty0 533, October 1986.
\newblock 14256.

\end{thebibliography}

\end{document}